\documentclass[10pt,twocolumn,letterpaper]{article}

\usepackage[pagenumbers]{cvpr} 

\usepackage{graphicx}
\usepackage{amsmath}
\usepackage{amssymb}
\usepackage{booktabs}

\usepackage[pagebackref,breaklinks,colorlinks]{hyperref}

\usepackage[capitalize]{cleveref}
\crefname{section}{Sec.}{Secs.}
\Crefname{section}{Section}{Sections}
\Crefname{table}{Table}{Tables}
\crefname{table}{Tab.}{Tabs.}

\def\cvprPaperID{*****} 
\def\confName{CS 4644}
\def\confYear{2022}

\begin{document}

\title{Efficient 3D Object Reconstruction using Visual Transformers}

\author{Rohan Agarwal\\
{\tt\small roaga@gatech.edu}
\and
Wei Zhou\\
{\tt\small wzhou322@gatech.edu}
\and
Xiaofeng Wu\\
{\tt\small xwu414@gatech.edu}
\and
Yuhan Li\\
{\tt\small yli3326@gatech.edu}
\and
Georgia Institute of Technology\\
North Ave NW, Atlanta, GA 30332\\
}
\maketitle


\section{Abstract}

Reconstructing a 3D object from a 2D image is a well-researched vision problem, with many kinds of deep learning techniques having been tried. Most commonly, 3D convolutional approaches are used, though previous work has shown state-of-the-art methods using 2D convolutions that are also significantly more efficient to train. With the recent rise of transformers for vision tasks, often outperforming convolutional methods, along with some earlier attempts to use transformers for 3D object reconstruction, we set out to use visual transformers in place of convolutions in existing efficient, high-performing techniques for 3D object reconstruction in order to achieve superior results on the task. Using a transformer-based encoder and decoder to predict 3D structure from 2D images, we achieve accuracy similar or superior to the baseline approach. This study serves as evidence for the potential of visual transformers in the task of 3D object reconstruction.

\section{Introduction}
\label{sec:intro}

3D reconstruction from 2D image is a process of transforming a 2D image into a 3D model represented by a point cloud, voxel mesh, or polygon mesh. It has been a challenging task, yet beneficial for many technologies and industries such as computer vision, animation, medical imaging, video game design, virtual reality and other computer graphics related industries. Improvements in accuracy and efficiency would be highly valuable.

The typical approach for 3D object reconstruction involves using 3D convolutional neural networks. However, these methods are computationally inefficient, and superior accuracy and efficiency has been achieved through using 2D convolution [2]. Meanwhile, in the field of computer vision in general, visual transformers have become increasingly popular, and on many vision tasks, outperform traditional convolutional architectures [1]. 

Thus, we set out to combine the benefits of 2D convolutional approaches on the task of 3D object reconstruction with the recent improvements of visual transformers. We attempt to enable superior performance and efficiency for 3D object reconstruction to make it more useful and practical in the numerous applications discussed above.

\section{Related Works}
\subsection{3D Object Reconstruction}

Previous works have tried many approaches for 3D object reconstruction. Typically, just as 2D convolutions are used in most image-based tasks, 3D convolutions are used in some form for this task. However, this requires coarser voxel representations of objects (similar to pixels in 2D) and is generally inefficient in both time and memory [2]. It has been demonstrated by Lin, Kong, and Lucey (2018) that 2D convolutions can more efficiently achieve state-of-the-art results for this task [2]. Their work predicted the 3D structure of an object using images from multiple viewpoints, generated through a 2D-convolution-based structure generator. They also presented a more efficient optimization method called "pseudo-rendering," which optimizes using 2D projection error. This approach also allows using point cloud representations instead of voxel representations.

There are other methods, such as those taking advantage of depth prediction, as presented by Yin, et. al. (2021) [3]. Their work trains a depth prediction model and a point cloud generation model separately. After combining the two in a single pipeline, the approach is able to outperform prior methods in 3D scene reconstruction.

\subsection{Visual Transformers}

As discussed by Lahoud, et. al. (2022), visual transformers are creating a new state-of-the-art in some vision tasks, both 2D and 3D, just as they have done in language [1]. They often outperform convolution-based approaches. There is also significant previous work on using transformers on both point-cloud and voxel representations (in addition to 2D pixel images), for object classification, object detection, segmentation, point cloud completion, pose estimation, and even 3D reconstruction to some extent. 

There are mainly two design principles related to building visual transformers. One principle is to get rid of convolutions entirely and replace it with self attention, like ViT [4] which embeds each 16*16 patch of the image as a vector and uses the encoder from the original transformer [5]. Another principle is to design a hybrid model with both convolutional layer and self attention like in MaxViT [6]. We have tried both architecture designs for our encoder and we will provide more detail in section 4.3.1.

\section{Methods}

Due to its efficiency and impressive results, and a focus on object reconstruction instead of scene reconstruction, we mostly model our approach after Lin, Kong, and Lucey (2018) [2]. The model uses a convolution encoder and decoder for feature extraction and viewpoint generation. This resembles the architecture of visual transformers, which we wish to replace the convolutional sections of the model with due to their potential to lead to superior results as shown by recent research in the field.

Our model consists of a visual transformer for converting an image to 8 images with depth and mask information at 8 different viewpoints. The projections are used to form a point cloud. The point cloud generation is then rendered to different projections at novel viewpoints and are compared with the ground truth projections to obtain loss. It is an efficient and reasonable loss function as we believe a good 3D reconstruction should produce a similar projection in different viewpoints compared to the ground truth.

\subsection{Data}
We used the dataset provided by Lin, Kong, and Lucey (2018) [2] at the following website: https://chenhsuanlin.bitbucket.io/3D-point-cloud-generation/. This dataset contains around 3000 data points of RGB images, depth information, and mask information. It also contains several categories of common objects, such as chairs. The depth and mask information, used as the ground truth information, was generated by rendering the corresponding CAD models at random viewpoints. Each image is 128x128 pixels.

\subsection{Preliminary Methods}

\subsubsection{Transformer-Based Structure Generator}
\begin{figure}[h]
\centering
\includegraphics[width=8cm]{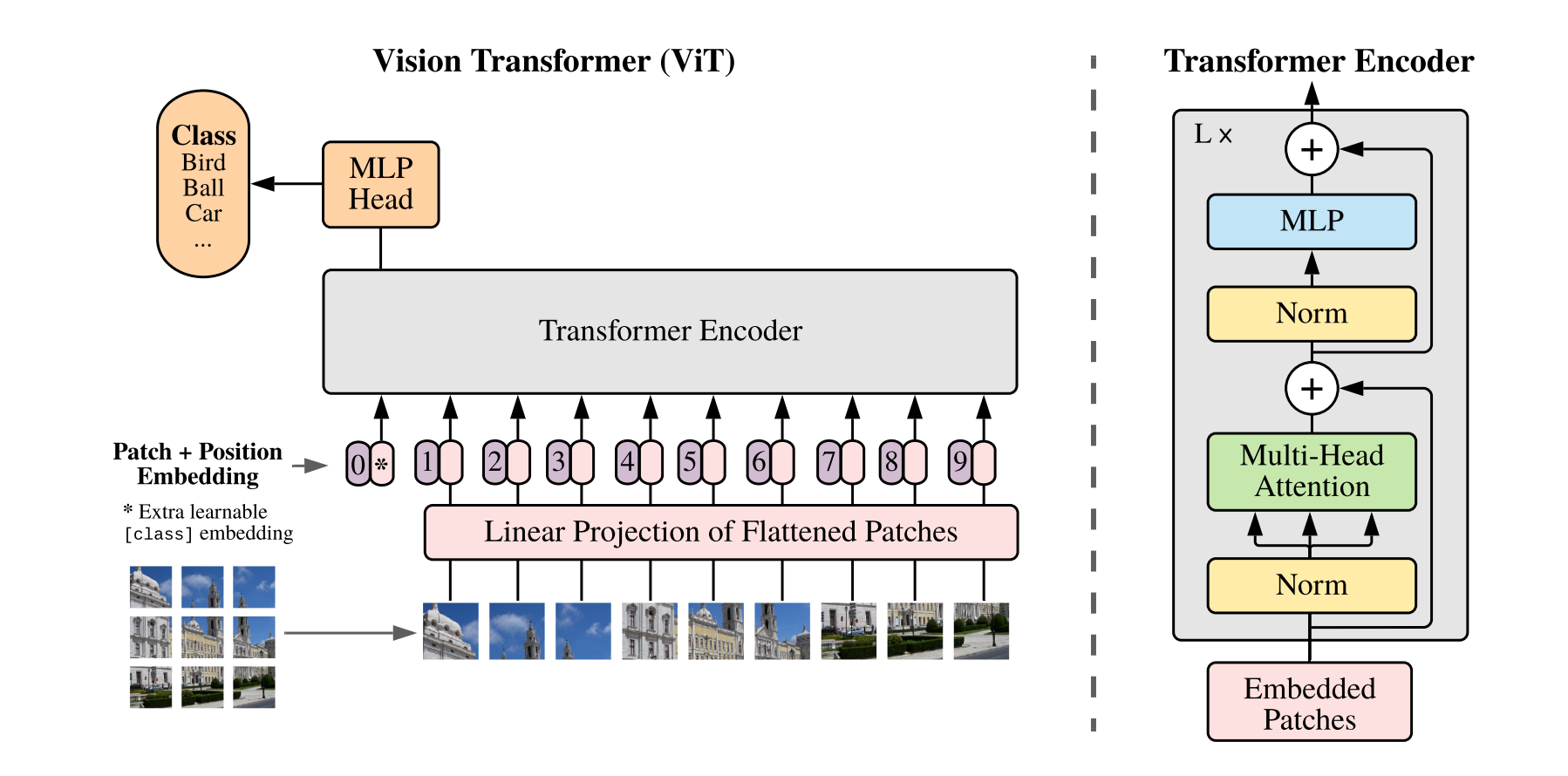}
\caption{Architecture of the original visual transformer and the transformer encoder used in it from the paper "An Image is Worth 16x16 Words: Transformers for Image Recognition at Scale" by Dosovitskiy et al. [4]}
\label{fig:vit}
\end{figure}

In the original structure generator, we replaced its encoder with a pretrained visual transformer \ref{fig:vit}. The outputs of the encoder are images with size 224$\times$224 pixels. To further shrink down the width and height, we passed it through two 2D convolutional blocks and five fully connected blocks. 

For the decoder part, we use four 2D deconvolutional blocks and one 2D convolutional block and bias term to upsample the image to the preferred size of 128$\times$128. The final output is an image with 16 channels, in which 8 channels represent the depth channel from 8 different viewpoints and the other 8 channels represent the mask from 8 different viewpoints representing whether each pixel should be kept during projection to 3D. Within each block, we add a batch normalization and a ReLU activation function to normalize the input for the next block and add nonlinearity to the model.

Another modification we have made is in the type of outputs from the model. The baseline model's output consisted of an XY coordinate matrix corresponding to depth information and mask. However, we noticed that the XY coordinate matrix is exactly the same for every image and thus we deemed it unnecessary to require the model to predict. Hence, our model only outputs depth information and mask. Thus, we deduced this might be why our model's result is much more stable than the baseline model when given the same image from different viewpoints as shown in Fig. \ref{fig:cmp}.

During training time, we froze all the parameters within the pretrained visual transformers except for the embedding layer and the final attention layer. The 2D convolutional and deconvolutional blocks as well as the fully connected blocks are all set to trainable. The training time of our model is much faster than the baseline model since our model consists of 894,128 trainable parameters while the baseline model consists of 28 million trainable parameters.

\subsubsection{Optimization}
The optimization method we are using is same one as used by Lin, Kong, and Lucey (2018) [2]. To avoid computationally expensive and intractable direct distance calculations between the 3D point cloud and ground truth label, we are using
joint 2D projection with pseudo-rendering. To be specific, we project canonical 3D points gained after the structure generator to 8 different preset viewpoints using rigid 3D transformation with upsampling and max pooling on inverse-depth to solve collision (same pixel contains more than one point in the point cloud). This gives us 8 images, where each pixel contains information on whether any point cloud is projected onto it by a binary mask. Then the loss is calculated as the cross-entropy loss of the mask and element-wise $L_1$ loss of the depth between 8 images of the ground truth model and the same viewpoints' 8 images of point clouds. We find this part very useful in our training because it can greatly reduce the generated point clouds from 30833.049 average number of points to 19565.377 average number of points for our transformer-based structure generator.

\subsection{Final Approach}

\begin{figure}[h]
\centering
\includegraphics[width=8cm]{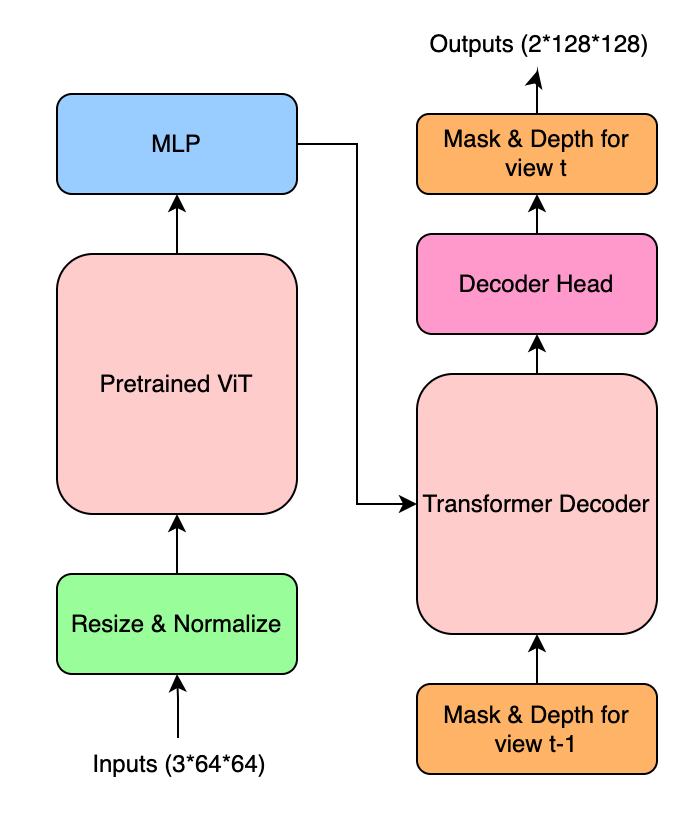}
\caption{Overview of the architecture of our new model.}
\label{fig:ourmodel}
\end{figure}

To improve results further, several further changes were made to the architecture from the preliminary method--mainly, replacing more convolutional parts of the architecture with more transformers (see Fig. \ref{fig:ourmodel}). 

\subsubsection{Encoder}
In theory, convolutional layer carries more inductive bias towards images and thus it's able to learn good features with less data compared to self attention layer. Therefore, we experimented using a hybrid design of convolutional layer and self attention layer like MaxViT described in [6], as well as Swin transformer described in [7]. However, after experiments, the original pretrained visual transformer is able to achieve the best results and thus we choose to use it as our model's encoder.
The pretrained visual transformer from [4] was kept as is for the encoder part of the structure generator, but the images were resized to 224$\times$224 pixels before being passed into the encoder. After the encoder, a MLP block is used to reduce the sequence length of the encoder's output from 179 to 64 so that it can be later interpreted in the decoder head as a feature map with height and width of 8.

\subsubsection{Decoder}
The most significant change from the preliminary method was replacing the decoder of the original structure generator with four layers of a transformer decoder layer in [5] with 30 million parameters, instead of the 2D convolutional layers from before. The transformer decoder at each timestep will generate a text-like output with shape batch size*sequence length*dimension. This is incompatible with the required output, which consists of depth and mask feature map with shape batch size*width*height.

Therefore, to address for this semantic gap, we add a decoder head with the target of correctly interpreting the transformer decoder's output as feature maps. During forward pass, the transformer decoder's output is reshaped from batch size*sequence length*dimension to batch size*channels*width*height, where dimension is interpreted as channels and width and height is the square root of sequence length. Our final design choice for the decoder output head is a 2D convolution with a 1$\times$1 kernel to reduce channels from 768 to 512 and a pixel shuffle layer that reshape a tensor of (channel*$r^2$,height,width) to (channel,height*$r$,width*$r$). This design choice is adopted from the visual transformer's decoder for masked image modeling. We also tried adding complexities to the decoder head by using layers of deconvolutional blocks, but there is no significant improvement in performance during experiments.

The output of the transformer decoder plus the decoder head should be a depth (1,128,128) and mask (1,128,128) matrix concatenated together (2,128,128) for each of the 8 points of view. So in total our model's output should be of shape (8,2,128,128). We believe that the depth and mask information from one point of view should be conditioned on the depth and mask information from the other points of view. Thus, the decoder's forward function is designed to be autoregressive. The decoder's hidden state is the encoder's last hidden state (after the linear block). The decoder’s output should be the depth and mask information for 8 different views, so at time $0$, we use a tensor of all ones as the input to signal the decoder to output the first view. At time $t$, we use the decoder’s output from time $t-1$ (i.e., the depth and mask info from view $t-1$) as inputs. We iterate 8 times to generate the output for all 8 views.

\subsubsection{Training Procedure}

To train our model, we froze the encoder's first six transformer layers (requires\_grad=False) and set its last six transformer layers to trainable. The linear block, decoder, and decoder head are all trained from scratch for a total of 76 million trainable parameters. 

We first trained for 80 epochs without joint 2D optimization to save time due to our limited computational resources. This can be done by comparing the outputted point cloud in eval mode with the ground truth point cloud for an error metric. The pretrained transformer is given a smaller learning rate since we should only have to finetune it. We use a cosine annealing learning rate scheduler.

After that, we further train the model with joint 2D optimization for 20 epochs. The learning rate is the same for the whole network and we again use a cosine annealing learning rate scheduler.

\section{Results}
\subsection{Preliminary Results}

We used the code provided at the same website [2], unmodified, to get baseline results. After making our modifications to the architecture we achieved the below preliminary results in Tables \ref{table:points}, \ref{table:3derror}.
The 3D error measures the average point-wise 3D Euclidean distance between two 3D point cloud models. This metric is defined bidirectionally as the distance from the predicted point cloud to the ground truth CAD model and vice versa.
\begin{table}[h!]
\centering
\begin{tabular}{||c c||} 
 \hline
 Method & Generated Points\\ [0.5ex] 
 \hline\hline
 Baseline & 18895.68 \\ 
 Our Method & 19565.38 \\
 \hline
\end{tabular}
\caption{The average number of 3D points generated by each method to form the point cloud after 20 training epochs. Our method is able to generate slightly more dense result than baseline.}
\label{table:points}
\end{table}

\begin{table}[h!]
\centering
\begin{tabular}{||c c c||} 
 \hline
 & 3D error metric & \\
 Method & Pred $\rightarrow$ GT & GT $\rightarrow$ Pred\\ [0.5ex] 
 \hline\hline
 Baseline & 5.682 & 7.214\\ 
 Our Method & 4.777 & 5.15\\
 \hline
\end{tabular}
\caption{The average 3D test error (scaled by 100) after 20 training epochs from both methods. Prediction to GT (ground truth) error measures 3D shape similarity and GT to prediction error
measures surface coverage. Our method outperforms the baseline method in both 3D shape similarity and surface coverage.}
\label{table:3derror}
\end{table}

\begin{figure}
\centering
\includegraphics[width=8cm]{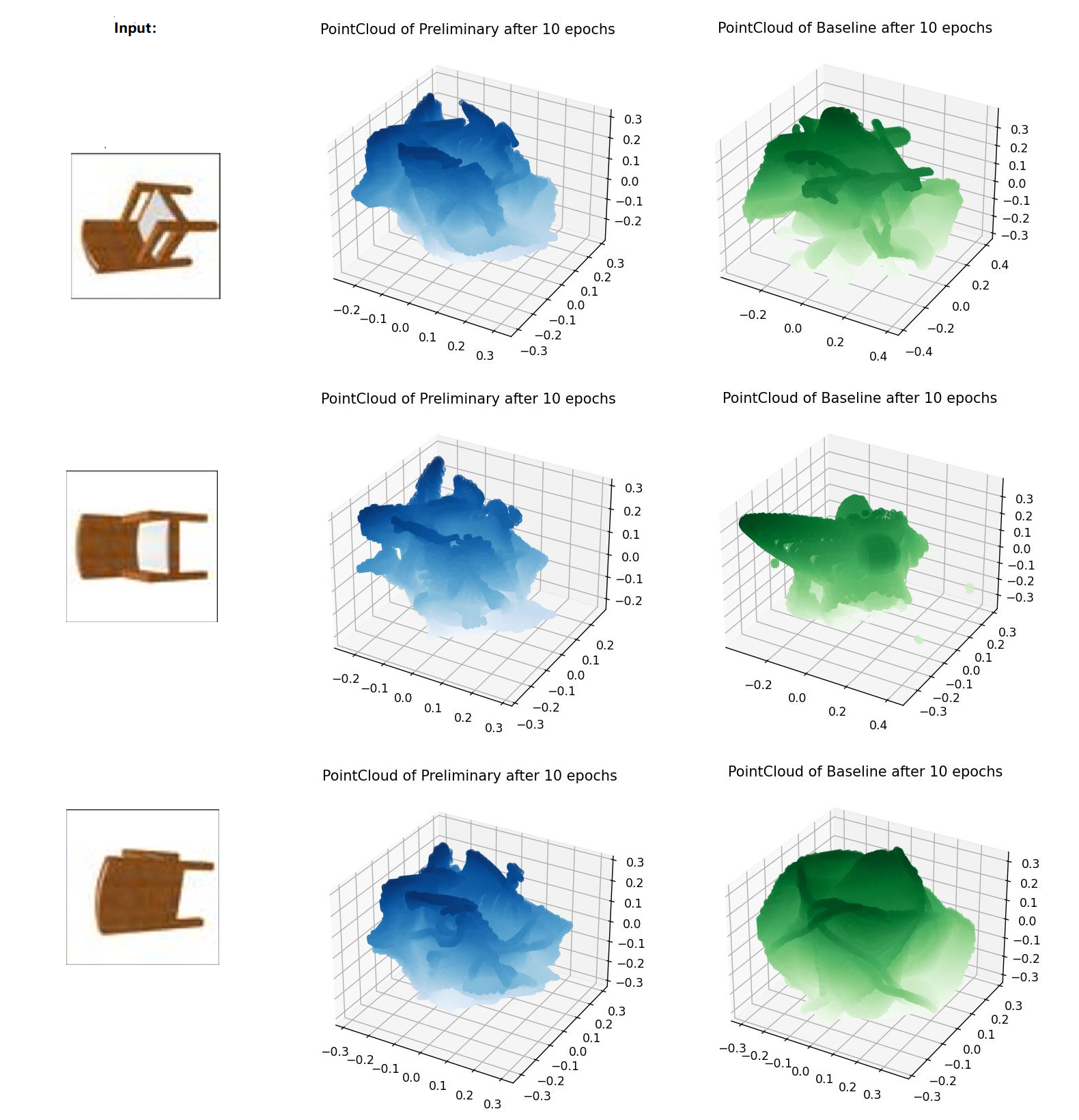}
\caption{This figure shows the 3D cloud point result by 10 epochs of our preliminary result and baseline result using the work by Lin, Kong, and Lucey (2018) [2]. The input is an image from one of multiples perspectives of the same chair. When giving an input of the back of the chair or when the chair only shows two of the four legs, our model shows a more stable result than the baseline.}
\label{fig:cmp}
\end{figure}

The visual comparison of the point clouds we produced is shown in Fig. \ref{fig:cmp}. While the result for the baseline model changed to a great extent when the perspective of images changed, the result of our preliminary model changes significantly less, which shows more robustness of our model. To be more specific, when the perspective of the image does not give too much information about the object's shape (when the back of the chair takes up most part of the image or when the front of the chair takes up most part of the image) it is difficult to get the depth information of the input. Compare to the preliminary model, the baseline one is more vulnerable when inputs are tricky. 
Another thing to notice: due to limited computing power, we only trained 20 epochs on both model's structure generator which resulted in a relatively poor point cloud of the model. However, the differences between two models are shown clearly.

\subsection{Final Results}

After the final changes to the model architecture and a more extensive training process with more epochs and across all 6678 objects provided in the dataset, as described in Section 4.3, we achieved new results, shown in Tables \ref{table:pointsfinalnoopt}, \ref{table:pointsfinalwopt}, \ref{table:3derrorfinalnoopt}, \ref{table:3derrorfinalwopt}.

\begin{table}[h!]
\centering
\begin{tabular}{||c c||} 
 \hline
 Method & Generated Points, No Opt.\\ [0.5ex] 
 \hline\hline
 Baseline & 32126.88 \\ 
 Our Method & 31219.99 \\
 \hline
\end{tabular}
\caption{The average number of 3D points generated by each method to form the point cloud after following the procedure described in Section 4.3.3, but before training with joint 2D optimization.}
\label{table:pointsfinalnoopt}
\end{table}

\begin{table}[h!]
\centering
\begin{tabular}{||c c||} 
 \hline
 Method & Generated Points, Opt.\\ [0.5ex] 
 \hline\hline
 Baseline & 25383.11\\ 
 Our Method & 15614.25  \\
 \hline
\end{tabular}
\caption{The average number of 3D points generated by each method to form the point cloud after following the procedure described in Section 4.3.3 completely.}
\label{table:pointsfinalwopt}
\end{table}

Looking at number of generated points, our method is effective at reducing that number compared to the baseline, especially when using joint 2D optimization.

\begin{table}[h!]
\centering
\begin{tabular}{||c c c||} 
 \hline
 & 3D Error Metric, No Opt. & \\
 Method & Pred $\rightarrow$ GT & GT $\rightarrow$ Pred\\ [0.5ex] 
 \hline\hline
 Baseline & 7.513 & 3.564 \\ 
 Our Method & 7.823 & 1.825 \\
 \hline
\end{tabular}
\caption{The average 3D test error (scaled by 100) after following the procedure described in Section 4.3.3, but before training with joint 2D optimization. Prediction to GT (ground truth) error measures 3D shape similarity and GT to prediction error
measures surface coverage.}
\label{table:3derrorfinalnoopt}
\end{table}

\begin{table}[h!]
\centering
\begin{tabular}{||c c c||} 
 \hline
 & 3D Error Metric, Opt. & \\
 Method & Pred $\rightarrow$ GT & GT $\rightarrow$ Pred\\ [0.5ex] 
 \hline\hline
 Baseline & 4.651 & 5.187 \\ 
 Our Method & 3.980 & 2.946 \\
 \hline
\end{tabular}
\caption{The average 3D test error (scaled by 100) after following the procedure described in Section 4.3.3 completely. Prediction to GT (ground truth) error measures 3D shape similarity and GT to prediction error measures surface coverage.}
\label{table:3derrorfinalwopt}
\end{table}

Looking at the errors, our method outperforms the baseline method, with the complete training method using joint 2D optimization. Without joint 2D optimization, our model is similar to or better than the baseline, depending on the metric. In some metrics, namely surface coverage, our model without joint 2D optimization outperforms the baseline even without joint 2D optimization, which would allow for significant savings in efficiency.

\begin{figure}
\centering
\includegraphics[width=8cm]{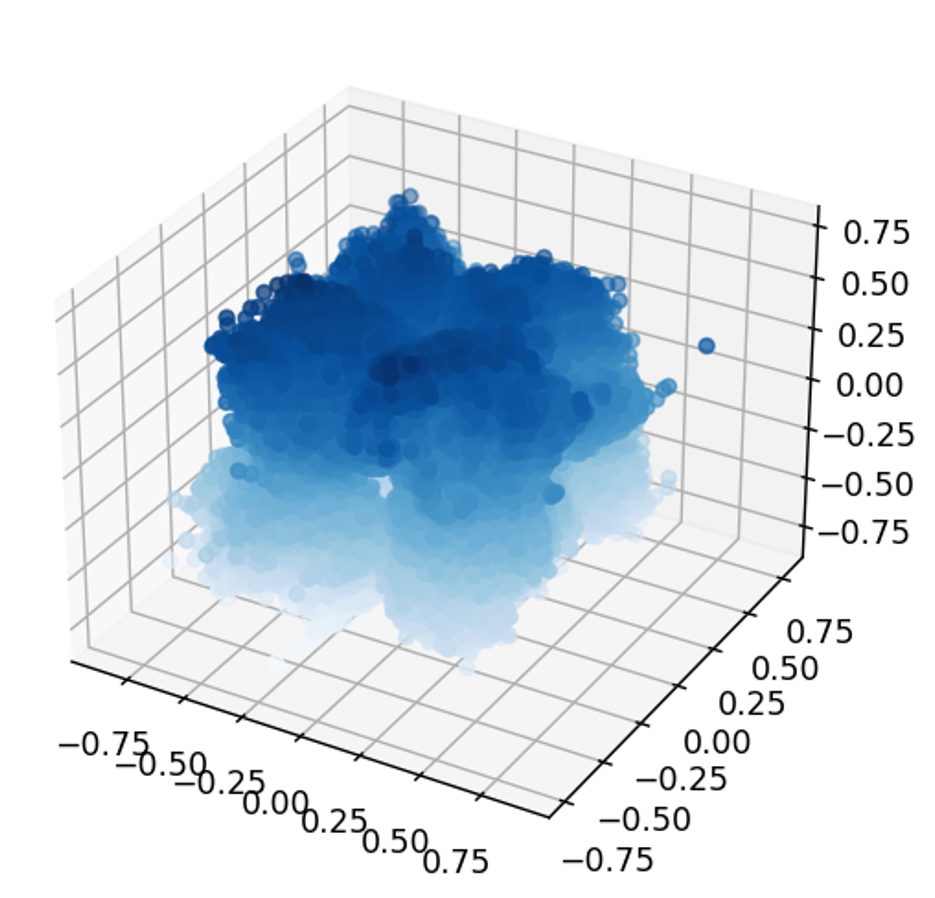}
\caption{This figure shows the 3D point cloud results from our method without joint 2D optimization.}
\label{fig:ournoopt}
\end{figure}

\begin{figure}
\centering
\includegraphics[width=8cm]{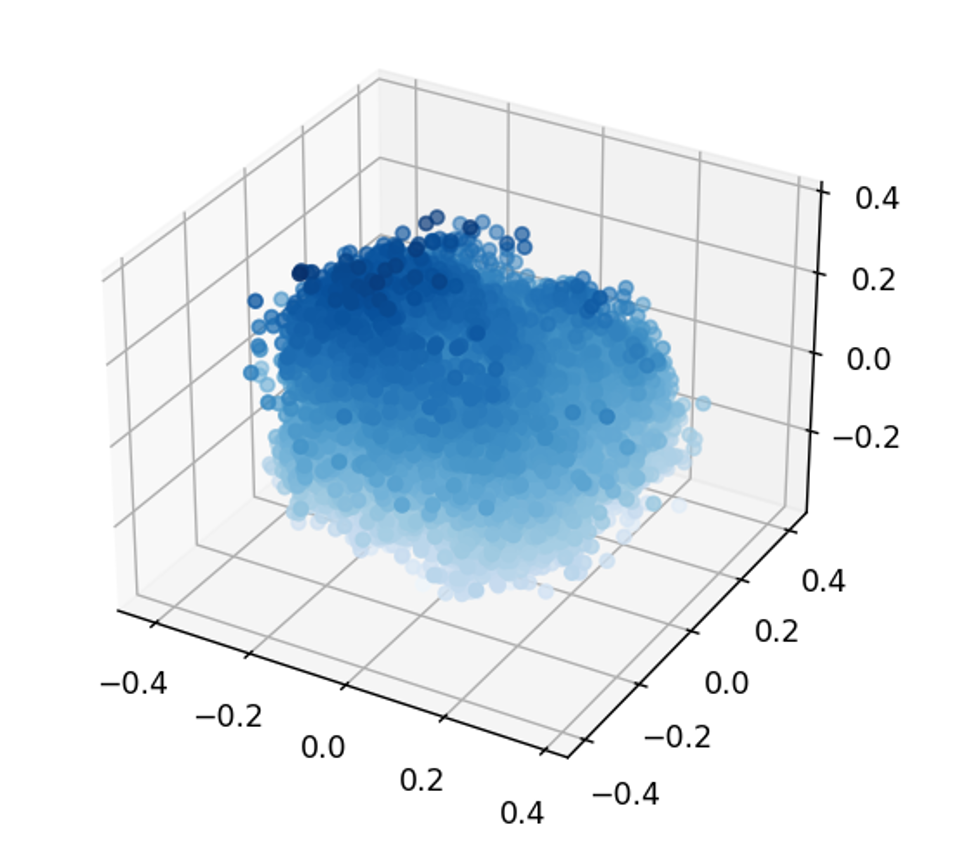}
\caption{This figure shows the 3D point cloud results from our method with joint 2D optimization.}
\label{fig:ourwopt}
\end{figure}

We present the outputted point clouds from the baseline and our model below in Figures \ref{fig:chairmodel}, \ref{fig:basenoopt}, \ref{fig:basewopt}, \ref{fig:ournoopt}, \ref{fig:ourwopt}. Neither model fully reconstructs the object, but that is to be expected due to the short training time of 80-100 epochs. This is a limitation of this study due to low computational resources. Regardless, it is qualitatively clear that both models are approximating the shape of the original model well, yet still produce very different outputs. It is interesting to consider how the choice of model changes the "approach" it takes to forming these point clouds over many iterations. 

Though the baseline (Figure \ref{fig:basewopt}) has a clearer "seat" than our method (Figure \ref{fig:ourwopt}), our method more closely approximates the squared shape of the original model in Fig. \ref{fig:chairmodel} than the baseline result. Without joint 2D optimization, our model (Fig. \ref{fig:ournoopt} is much closer to the correct shape of the chair than the baseline in Fig. \ref{fig:basenoopt}; this is an avenue for potential efficiency gains if joint 2D optimization could be fully eliminated in future work. Even still, it is hard to definitely make a comparison qualitatively due to the low training time. Yet along with the lower errors, it is evident that our model using primarily visual transformers outperforms or at least matches the baseline model that uses primarily 2D convolution. 

\begin{figure}
\centering
\includegraphics[width=8cm]{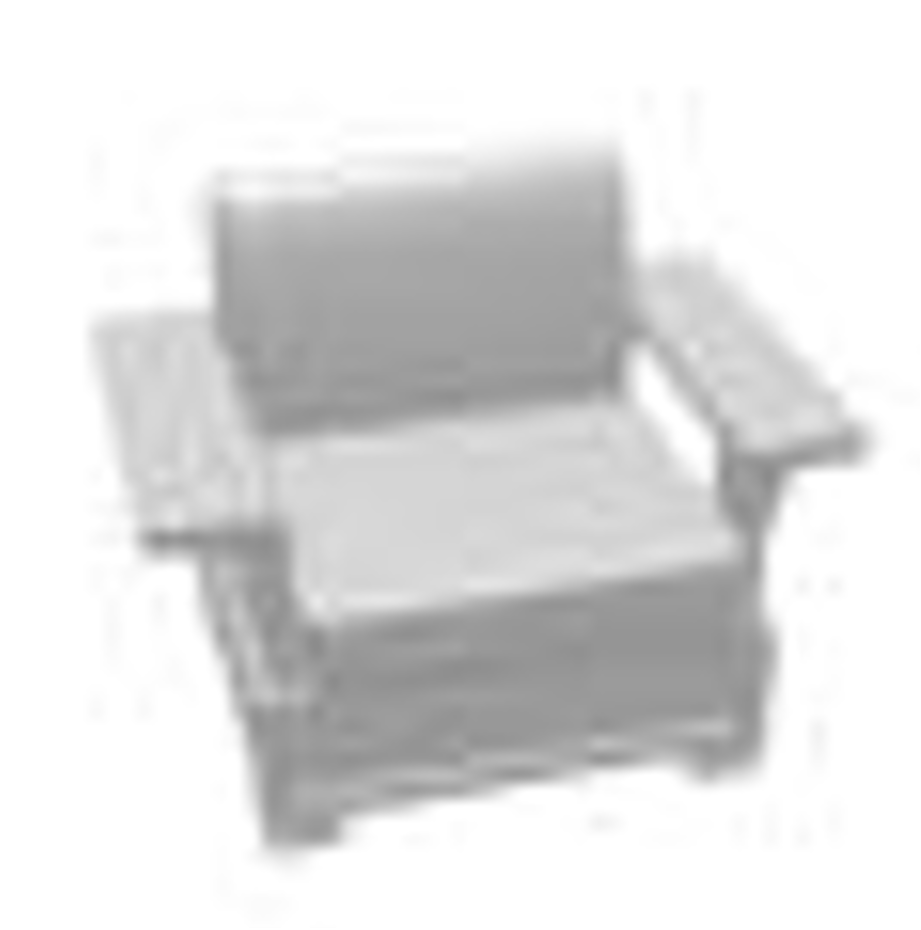}
\caption{This figure shows the ground truth model given to get outputs shown in Figures \ref{fig:basenoopt}, \ref{fig:basewopt}, \ref{fig:ournoopt}, \ref{fig:ourwopt}.}
\label{fig:chairmodel}
\end{figure}

\begin{figure}
\centering
\includegraphics[width=8cm]{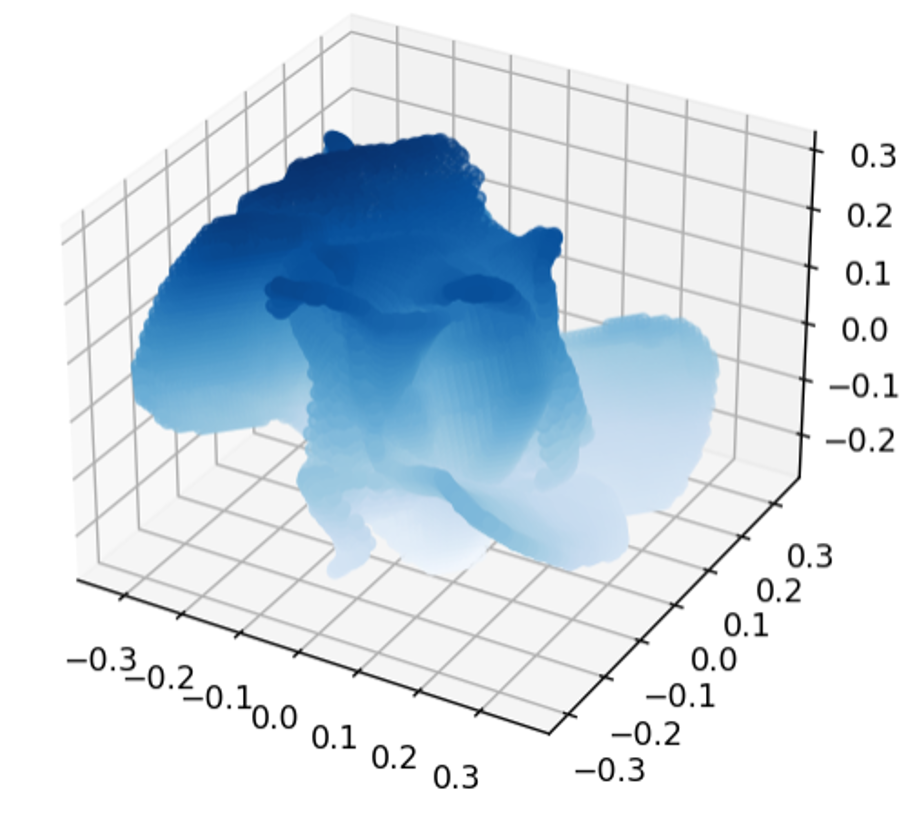}
\caption{This figure shows the 3D point cloud results from the baseline method without joint 2D optimization.}
\label{fig:basenoopt}
\end{figure}

\begin{figure}
\centering
\includegraphics[width=8cm]{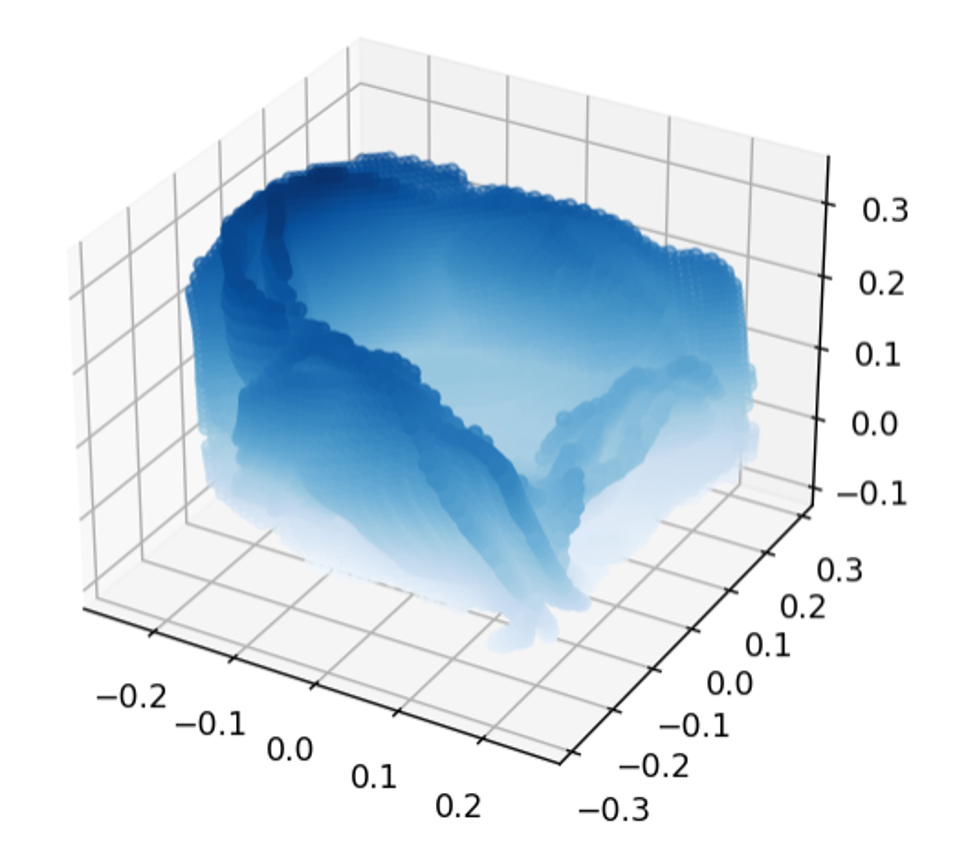}
\caption{This figure shows the 3D point cloud results from the baseline method with joint 2D optimization.}
\label{fig:basewopt}
\end{figure}

\begin{figure}
\centering
\includegraphics[width=8cm]{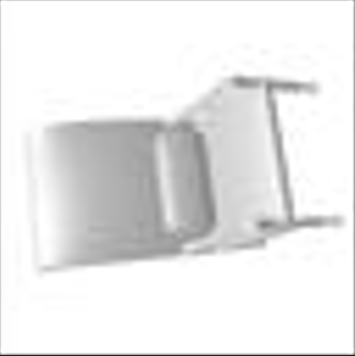}
\includegraphics[width=8cm]{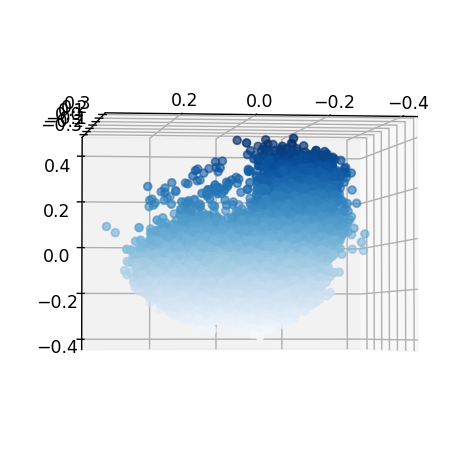}
\caption{This figure shows our model's output on another object.}
\label{fig:example1ours}
\end{figure}

\begin{figure}
\centering
\includegraphics[width=8cm]{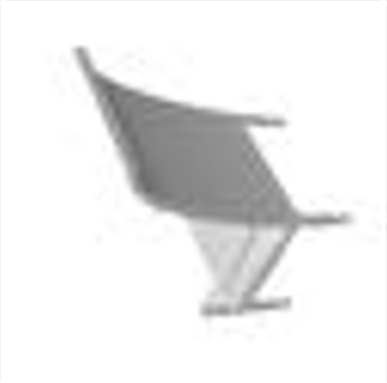}
\includegraphics[width=8cm]{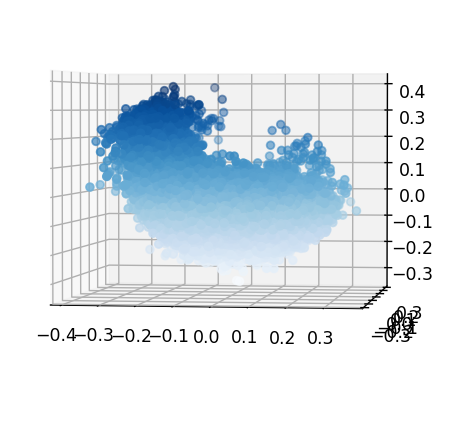}
\caption{This figure shows our model's output on another object.}
\label{fig:example2ours}
\end{figure}

\begin{figure}
\centering
\includegraphics[width=8cm]{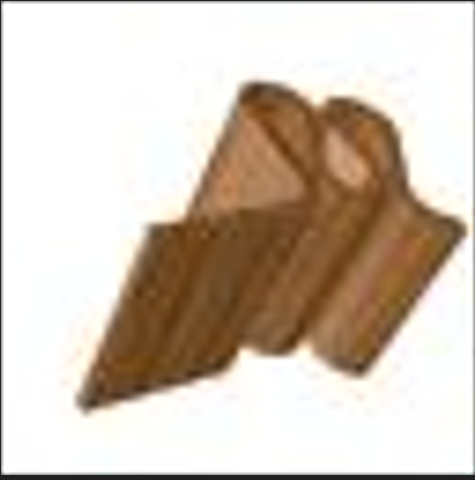}
\includegraphics[width=8cm]{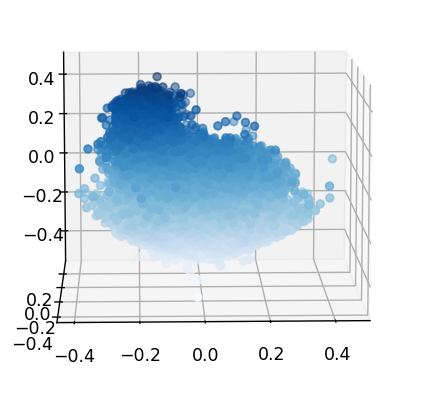}
\caption{This figure shows our model's output on another object.}
\label{fig:example3ours}
\end{figure}

Below we present several more examples of point cloud outputs from our model on other 3D objects that demonstrate its ability to approximate the shape, despite a short training process (Figures \ref{fig:example1ours}, \ref{fig:example2ours}, \ref{fig:example3ours}).

\section{Conclusion}

Overall, this study shows the potential of visual transformers in accomplishing efficient 3D object reconstruction. By taking an approach to this task with state-of-the-art results and high efficiency [2], and replacing the convolutional portions of the technique with visual transformers, we are able to achieve similar results in all cases and better in some cases. While we provide a new potential method for 3D object reconstruction applications, more importantly to this area of research, this work provides evidence that visual transformers can effectively tackle the problem of 3D object reconstruction from a single image. It shows that visual transformers should be explored further for this specific task of 3D object reconstructions, other variations of the task (such as starting with images from multiple viewpoints), and more generally, for other 2D and 3D vision tasks.

Due to limited computational resources, this work should ideally be repeated with more time so that we can make more thorough comparisons and more interesting qualitative comparisons (as the generated point clouds were not very precise for either method with such short training time). The original paper for our baseline method used 1000 epochs to generate results [2], but we were only able to train for 100 epochs. Another limit we found was how our model addresses 3D geometric information. Based on our results, our model can generate 3D shapes with high surface coverage but low 3D similarity to the ground truth. We attempted developing loss function using point-wise distance between vertices of polygons and generated point cloud, which is more efficient than using two point clouds, but our hardware memory was a limit and requires much more tuning when implement it. Thus, it is worth keep trying to incorporate more 3D geometric information when computing loss in the future. Other future work could explore more custom-designed transformer architectures for improvements. It should explore how visual transformers can improve other approaches to this problem, such as those more popular techniques based on 3D convolution (especially since visual transformers have been used directly on 3D before [1]), depth recognition, and more.






{\small
\bibliographystyle{ieee_fullname}
\bibliography{egbib}
}
[1] J. Lahoud, J. Cao, F. S. Khan, H. Cholakkal, R. M. Anwer, S. Khan, and M.-H. Yang, “3D vision with transformers: A survey,” arXiv.org, 08-Aug-2022. [Online]. Available: https://arxiv.org/abs/2208.04309. [Accessed: 31-Oct-2022]. 

[2] C.-H. Lin, C. Kong, and S. Lucey, “Learning efficient point cloud generation for dense 3D object reconstruction,” Proceedings of the AAAI Conference on Artificial Intelligence, vol. 32, no. 1, 2018. 

[3] W. Yin, J. Zhang, O. Wang, S. Niklaus, L. Mai, S. Chen, and C. Shen, “Learning to recover 3D scene shape from a single image,” 2021 IEEE/CVF Conference on Computer Vision and Pattern Recognition (CVPR), 2021. 

[4]  A. Dosovitskiy, L. Beyer, A. Kolesnikov., et al. 2021. "An image is worth 16x16 words: Transformers for image recognition at scale." arXiv.org. [Online]. Available: https://arxiv.org/abs/2010.11929v2. [Accessed: 30-Nov-2022].

[5] A. Vaswani, N. Shazeer, N. Parmar, J. Uszkoreit, L. Jones, A. N. Gomez, L. Kaiser, and I. Polosukhin, “Attention is all you need,” arXiv.org, 06-Dec-2017. [Online]. Available: https://arxiv.org/abs/1706.03762. [Accessed: 05-Dec-2022]. 

[6] Z. Tu, H. Talebi, H. Zhang, F. Yang, P. Milanfar, A. Bovik, and Y. Li, “MaxViT: Multi-axis vision transformer,” arXiv.org, 09-Sep-2022. [Online]. Available:https://arxiv.org/abs/2204.01697. [Accessed: 05-Dec-2022]. 

[7] Z. Liu, Y. Lin, Y. Cao, H. Hu, Y. Wei, Z. Zhang, S. Lin, and B. Guo, “Swin Transformer: Hierarchical vision transformer using shifted windows,” arXiv.org, 17-Aug-2021. [Online]. Available: https://arxiv.org/abs/2103.14030. [Accessed: 05-Dec-2022]. 

\end{document}